\documentclass[10pt,twocolumn,letterpaper]{article}

\usepackage{iccv}
\usepackage{times}
\usepackage{epsfig}
\usepackage{graphicx}
\usepackage{amsmath}
\usepackage{amssymb}

\usepackage{graphicx}
\usepackage{amsmath}
\usepackage{amssymb}
\usepackage{booktabs}
\usepackage{algorithm}
\usepackage{algorithmic}
\usepackage{pifont}
\newcommand{\cmark}{\ding{51}}%
\newcommand\xmark{\ding{55}}%
\usepackage[breaklinks=true,bookmarks=false]{hyperref}

\iccvfinalcopy 

\def\iccvPaperID{****} 

\ificcvfinal\fi

\begin{document}

\title{Weakly-supervised 3D Pose Transfer with Keypoints}

\author{
Jinnan Chen \qquad Chen Li \qquad Gim Hee Lee  \\
Department of Computer Science, National University of Singapore\\
{\tt\small \{jinnan.c, lichen\}@u.nus.edu \qquad gimhee.lee@nus.edu.sg}
}


\maketitle
\ificcvfinal\thispagestyle{empty}\fi

\begin{abstract}
The main challenges of 3D pose transfer are: 1) Lack of paired training data with different characters performing the same pose; 2) Disentangling pose and shape information from the target mesh; 3) Difficulty in applying to meshes with different topologies.
We thus propose a novel weakly-supervised keypoint-based framework to overcome these difficulties. Specifically, we use a topology-agnostic keypoint detector with inverse kinematics to compute transformations between the source and target meshes. Our method only requires supervision on the keypoints, can be applied to meshes with different topologies and is shape-invariant for the target which allows extraction of pose-only information from the target meshes without transferring shape information. We further design a cycle reconstruction to perform self-supervised pose transfer without the need for ground truth deformed mesh with the same pose and shape as the target and source, respectively.
We evaluate our approach on benchmark human and animal datasets, where we achieve superior performance compared to the state-of-the-art unsupervised approaches and even comparable performance with the fully supervised approaches. We test on the more challenging Mixamo dataset to verify our approach's ability in handling meshes with different topologies and complex clothes. Cross-dataset evaluation further shows the strong generalization ability of our approach. Our source code is available at: \url{https://github.com/jinnan-chen/3D-Pose-Transfer}.
\end{abstract}

\begin{figure}[t] 
\centering 
\includegraphics[width=0.48\textwidth]{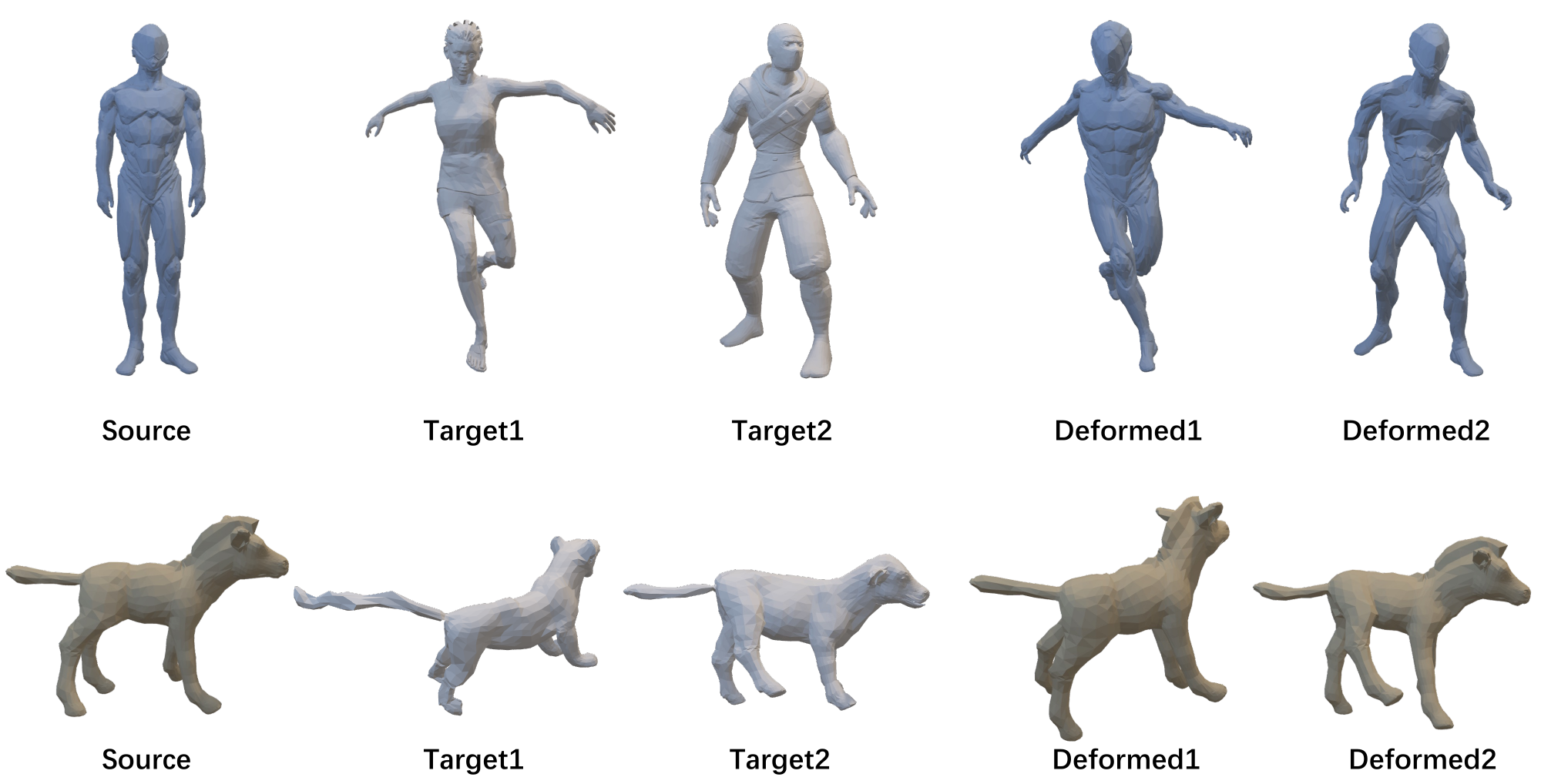} 
\caption{Examples of our pose transfer results on human and animal. Given a source mesh and a target mesh, we aim to transfer the pose from the target to the source. 
} 
\label{fig:teaser} 
\end{figure}

\section{Introduction}
\label{intro}
3D Pose transfer refers to transferring the pose from a target input to a source input while keeping the identity information of the source at the same time.
Pose transfer is an important research topic in computer vision because of its wide applications in many real-world applications such as augmented/virtual reality (AR/VR), movie making, gaming, metaverse, etc. 

Significant progress has been made for 3D pose transfer with the development of deep learning-based methods \cite{disentangle20, 3dpt21, neuralpt20, geotrans21, ieugan21}. However, 3D pose transfer remains a very challenging task due to the lack of paired training data 
since it is difficult to obtain 
data of two characters performing the same pose. To alleviate this problem, existing works \cite{neuralpt20,3dpt21,geotrans21} generate such paired data synthetically
using the SMPL \cite{smpl15} and SMAL \cite{smal17} models. 
The advantage of synthetic data is that it is convenient to generate paired training data by keeping the pose parameter of different meshes the same. 
However, the data generated from SMPL and SMAL has a strong bias for both shape and pose information due to the model parameterization. Consequently, a network trained with synthetic data cannot adapt well to real 3D meshes with large shape and pose variations. 
Unsupervised approaches are proposed in \cite{disentangle20,ieugan21,limp20} to circumvent the requirement for paired training data. These works adopt an auto-encoder-based framework to learn the shape and pose embeddings implicitly. 
The pose transfer can then be achieved by swapping the pose code between the source and target meshes. Although data-efficient, these works only show results for 3D pose transfer between meshes with the same topology. Furthermore, the shape and pose information are not fully disentangled in \cite{limp20,disentangle20} 
due to their implicit representation. 

We propose a 3D pose transfer model weakly-supervised with keypoints to mitigate the limitations of existing works. 
Our method is \textit{weakly-supervised} since we only need the supervision on keypoints instead of ground truth deformed mesh. As shown in Fig.~\ref{fig:teaser}, our approach achieves accurate 3D pose transfer although we do not use ground truth paired data. 
Specifically, we first detect keypoints on both source and target meshes with a topology-agnostic Pointnet \cite{pointnet17}. We then compute the transformation matrices between the two sets of keypoints with the differentiable Scalable Inverse and Forward Kinematics (IK/FK) functions, and propagate the transformations to all vertices of the source mesh with Linear Blending Skin (LBS)-based motion propagation. 
To circumvent the lack of the ground truth LBS skinning weights, we also design a Gaussian Mixture Model (GMM)-based pseudo label to supervise
the skinning weights.
We choose 
keypoints because it is easy to detect and there are correspondences between subjects of the same category. Ground truth for keypoints is also available for most datasets. Furthermore, in contrast to the implicit methods, our combination of keypoint-based transformation estimation and differentiable IK/FK helps \textit{disentanglement} of the pose from the shape information of the target.

Given that direct supervision for the deformed mesh is not available, we propose a cycle reconstruction that can be trained on realistic stylized meshes without the ground truth deformed mesh: the deformed mesh is exploited as a new target pose to reconstruct the original target mesh. This cycle reconstruction 
enforces the pose 
transfer from the target to the source mesh. 
Since our model operates on 
keypoints, it is \textit{topology-agnostic} and thus can be applied to meshes with large shape variations and different topologies.
Furthermore, shape regularizers are also added to enforce the consistency between the deformed and source meshes. 
We evaluate our approach on the commonly used SMPL-synthetic dataset NPT~\cite{neuralpt20} , SMAL-based~\cite{smal17} animal dataset and FAUST \cite{faust14} from real scans, where we outperform existing unsupervised approaches and even achieve comparable performance with fully-supervised approaches. To further evaluate our method on more complex and diverse topologies, we also collect a new 3D mesh dataset from Mixamo \cite{mixamo22}, where we show better performance than the existing work.
Experiments show the superiority of the proposed method compared to the state-of-the-art 3D pose transfer methods. 

\textbf{Our contributions} can be summarized as: 1) We propose a new 3D pose transfer framework for training data without ground truth supervision on the output deformed mesh. 2) Our approach is the first keypoint-based data-driven method for 3D pose transfer, and achieves better shape and pose disentanglement when combined with IK/FK. 3) Our approach is topology-agnostic, and thus can be applied to meshes with different topologies and non-T-pose source mesh. 4) We achieve superior performance compared to the state-of-the-art unsupervised approaches on FAUST dataset and supervised approaches on Mixamo dataset, and even achieve comparable performance with the fully supervised approaches on the NPT and SMAL datasets.

\section{Related work}
\paragraph{Fully-supervised 3D pose transfer.}
3D pose transfer, also known as deformation transfer has been intensively studied in both computer vision and graphics for a long time. DT~\cite{dt04} is a traditional explicit deformation transfer method for unregistered mesh with different topologies. It requires keypoints annotation and input meshes in the T-pose for optimization which is not always available. 
%
%
Recently, some deep learning-based 3D pose transfer methods have been proposed~\cite{neuralpt20, 3dpt21, geotrans21}. These works have achieved promising pose transfer performance by merging source and target information in an implicit way with paired ground truth supervision. NPT~\cite{neuralpt20} is the first end-to-end 3D pose transfer work. They treat the 3D pose transfer as a style transfer problem extended from \cite{ain17} to the point cloud domain with the content as the identity information and the pose as the style information.
3DPT \cite{3dpt21} is then proposed to directly build the correspondence between source and target vertices by solving an optimal transport problem with the Sinkhorn-Knopp algorithm. Based on the estimated correspondence matrix, a coarse deformed mesh is obtained. 

Inspired by \cite{elaimtrans19} for image style transfer, an elastic instance normalization (ElaIN) module is further proposed to blend the statistics of the original features (coarse results) and the learned parameters from external data (target) elastically. GCT \cite{geotrans21} uses Transformer \cite{trans17} as the more powerful backbone for feature extraction. Furthermore, a direction-aware central geodesic contrastive loss is added to minimize the geodesic features for all the edges between the deformed and the ground truth meshes. However, all of these approaches require strong supervision with paired data, which is hard to obtain in practice. In contrast, we design a weakly-supervised 3D pose transfer approach, which only requires keypoints for supervision. More recently, SKF \cite{sk22} designs a pose transfer framework, which is skeleton-free and able to handle meshes with different topologies. However, T-pose for both source and target meshes is required even during the inference, which is not available for most cases.
On the contrary, our approach can transfer pose from the target mesh to the source mesh with any pose without the requirement of a T-pose shape.

\vspace{-3mm}
\paragraph{Unsupervised 3D pose transfer.}
There are also unsupervised 3D mesh disentangling methods \cite{disentangle20,limp20,neuralmorph21,ieugan21}. These works seek to use an auto-encoder structure to implicitly convert the input mesh into shape and pose latent code without the ground truth paired data as supervision. SPD \cite{disentangle20} designs a novel framework for unsupervised shape and pose disentanglement with latent codes, which can also be applied to 3D pose transfer by swapping the latent code between the source and target mesh. Furthermore, an as-rigid-as-possible constraint is added to the generated meshes with the source meshes to keep the shape information. However, the limitation of this work is that fixed mesh topology is a necessity for the spiral CNN network structure, thus limiting the model generalization ability. LIMP~\cite{limp20} adopts metric preservation to control the amount of geometric distortion incurring in the latent space and a differentiable geodesic loss for intrinsic preservation. More recently, an unsupervised method is proposed for 3D pose transfer in ~\cite{unsupervised22} with cross consistency and dual reconstruction.
In comparison to these methods, our approach is topology-agnostic, where we can handle meshes with different topologies and 
disjoint parts. Moreover, our keypoint-based motion representation and propagation help to better disentangle the pose from the shape information during pose transfer.
\vspace{-3mm}
\paragraph{Keypoints-based deformation.}
Keypoints-based deformation are well-studied in \cite{mvc05,kd21,cage20,handle21}. They achieve high-quality deformation based on keypoints detection and skinning-based motion propagation. The 3d keypoints are detected with the prior from Farthest point sampling in an unsupervised way \cite{kd21} or with pre-processed optimization to find the closest cages \cite{cage20}. For shape-preserving, cage-based \cite{mvc05,cage20} methods use Mean Value Coordinates (MVC) for smooth deformations. Other deformation methods adopt as-rigid-as-possible \cite{arap07} or Laplacian \cite{meshpro06} regularization to preserve shape details \cite{cage20}. However, their focus is on rigid objects without any articulated motion. In comparison, we are the first to apply keypoints-based deformation on the articulated objects for pose transfer.

\vspace{-3mm}
\paragraph{Comparison.}
We show our advantages over other methods in Tab.~\ref{table:adv} in terms of: 1) requirement of ground truth mesh for supervision; 2) requirement of additional T-pose during inference; 3) ability to transfer across different topologies; 4) implicit or explicit disentanglement. Note that our method and \cite{sk22} use transformation matrices as an explicit disentanglement method to exclude shape information being transferred. Implicit methods simply output shape and pose code, thus information tends to entangle together.

\begin{table}[htbp]
\centering
\small
\setlength{\tabcolsep}{0.08cm}
\begin{tabular}{|c|c|c|c|c|}
\hline
\multicolumn{1}{|c|}{Method} & \multicolumn{1}{c|}{w/o GT}&\multicolumn {1}{c|}{w/o T-pose}&\multicolumn {1}{c|}{Cross topologies}& \multicolumn {1}{c|}{Disentanglement}\\
\hline
\cite{dt04}&\cmark &\xmark &\cmark  &Explicit\\
\cite{limp20}&\cmark &\cmark &\cmark  &Implicit\\
\cite{disentangle20}&\cmark &\cmark &\xmark  &Implicit\\
\cite{neuralpt20}&\xmark &\cmark &\xmark  &Implicit\\
\cite{3dpt21}&\xmark&\cmark &\xmark  &Implicit\\
\cite{geotrans21}&\xmark&\cmark &\cmark&Implicit\\
\cite{sk22}&\xmark&\xmark &\cmark &Explicit\\
Ours&\cmark &\cmark &\cmark  &Explicit\\
\hline
\end{tabular} \vspace{2mm}
\caption{Our advantages compared with existing methods.} \vspace{-2mm}
\label{table:adv}
\end{table}

\begin{figure*}[t] 
\centering 
\includegraphics[width=1\textwidth]{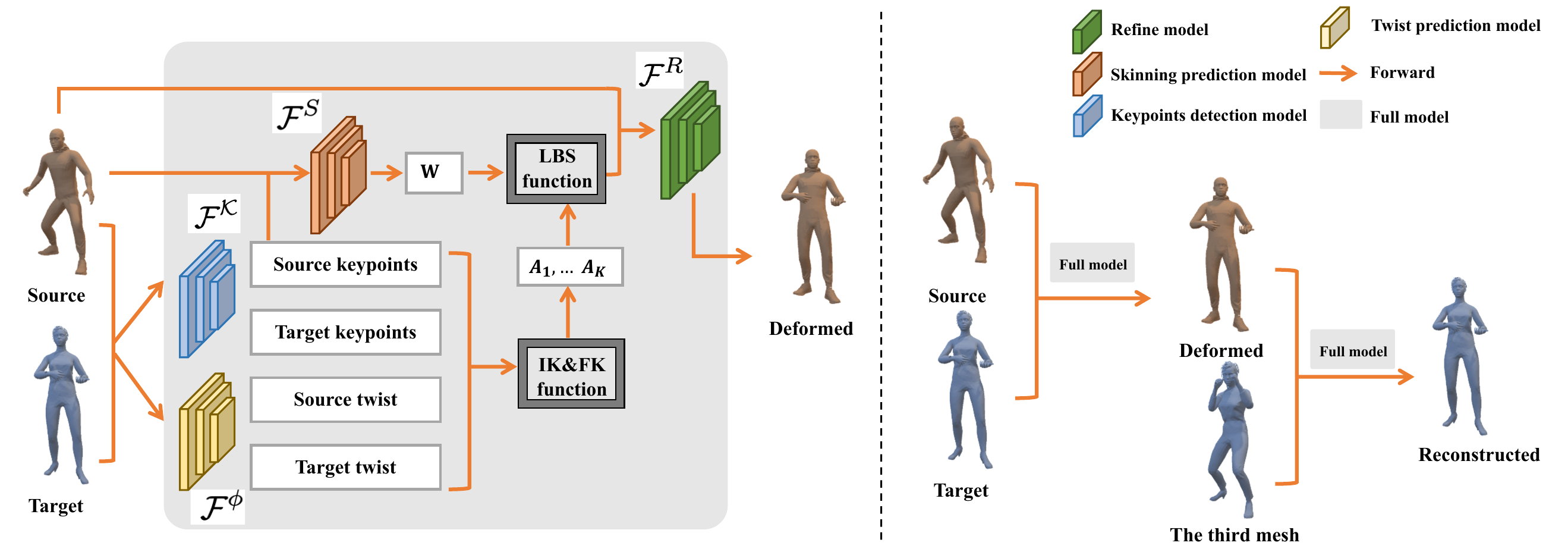} 
\caption{The overall framework of our proposed approach.  The left part is our pipeline for pose transfer, which contains four learnable components: a keypoints detection network, a twist prediction network, a skinning weights prediction network, and a refinement network, and two functions: an Inverse and Forward Kinematics function and an LBS function. The right part is an illustration of the cycle reconstruction process. The yellow and blue meshes represent two different characters.}
\label{fig:framework}
\end{figure*}

\section{Our Method}
\label{Chapter3}
\paragraph{Problem definition.} Let ${X_{p1,i1}}$ and ${X_{p2,i2}}$ be the source mesh 
and the target mesh, 
where $p$ represents the pose information and $i$ represents the identity information. 
The objective
is to generate 
a deformed
mesh ${X_{p2,i1}}$ with the identity information from the source mesh and the pose information from the target mesh. Formally, we aim to learn a general function $\mathcal F(\cdot)$, represented with a deep network, such that:  
\begin{equation}
\label{eq:obj}
\mathcal F(X_{p1,i1},X_{p2,i2}) \mapsto X_{p2,i1}.
\end{equation}
\paragraph{Overview.} The main challenge is to train the network without paired training data, where the ground truth for $X_{p2,i1}$ in Eqn.~\eqref{eq:obj} is not available. We thus introduce a new framework to circumvent this issue. As shown 
on the left 
of Fig.~\ref{fig:framework}, our proposed framework contains four learnable components: 1) a keypoints detection model; 2) a twist prediction model; 3) a skinning prediction model; 4) a refinement model, two non-learnable parts: the IK and FK functions, and an LBS function.
Specifically, our approach learns the 3D pose transfer task in five steps. 1) \textbf{Keypoints Detection.} We start with keypoint detection of the input source and target meshes. 2) \textbf{Scalable Inverse and Forward Kinematics.} We then estimate the relative rotation matrices between the source and target using scalable inverse kinematics based on the corresponding detected keypoints. Forward kinematics is also adopted to compute the global transformation matrix for each bone of the source mesh. 3) \textbf{Motion propagation with GMM-based LBS.} 
Subsequently, we propagate the transformation matrix of each keypoint to all vertices of the source mesh. Since there is no ground truth supervision for the skinning weights, we design a GMM module for pseudo labels. 4) \textbf{Mesh Refinement.} To model the non-linear deformations, we further add a refinement network to model the non-rigid deformation to recover fine-grained details.  5) \textbf{Cycle and Self Reconstruction.} Given that there is no direct supervision for the output, we introduce a cycle and a self reconstruction that can be self-supervised with the input meshes. 


\subsection{Keypoints Detection}
We first detect a set of keypoints for both the source and target meshes using a keypoint detector. 
Specifically, we define the keypoints as the joints in the SMPL model for the human shapes and SMAL model for the animal shapes such that the keypoints ground truth can be directly computed using the joint regressor \cite{smpl15}. For the non-template-based 3D meshes, \eg meshes in the Mixamo Dataset, we select the joints that are semantically similar to the SMPL keypoints with annotations in the dataset for keypoint supervision. 
To handle meshes of different topologies, we use a simple Pointnet and MLP as the keypoint detector, which we denote as $\mathcal{F^{K}}(\cdot)$. Given all the vertices of the source and target meshes, represented by $V_{s}$ and $V_{t}$, respectively, the keypoint detector predicts the keypoints as:
\begin{equation}
\centering
\begin{aligned}
k_{s}=\mathcal{F^{K}}(V_{s}), \quad k_{t}=\mathcal{F^{K}}(V_{t}).
\end{aligned} 
\label{kpnet} 
\end{equation}
The keypoint detector is supervised with the $L_2$ distance:
\begin{equation}
\centering
\begin{aligned}
\mathcal{L}_{k}=||k_{s}-&k_{s}^{gt}||_{2}+||k_{t}-k_{t}^{gt}||_{2},
\end{aligned} 
\label{kploss} 
\end{equation}
where $k^{gt}$ denotes ground truth keypoints.

\subsection{Scalable Inverse and Forward Kinematics}
Given the keypoints of the source and target meshes in different shapes, we aim to infer the relative motion between them. The most naive way is to directly subtract target keypoints from source keypoints as the motion representation. However, this 
lead to entanglement of the pose and shape information since different scales of the source and target meshes 
also contribute to the motion representation. Therefore, we represent the keypoint motion with a transformation matrix instead of the motion vector. 
Different from the case of the original inverse kinematic \cite{comput85,robo84,leastik05} where the shape keeps fixed, the shape of the source and target meshes are generally different in our case, \eg the bone length. In view of this, we introduce a scalable IK to compute the relative rotations suitable for source and target with different shapes.
Particularly, we sequentially compute the relative rotation matrices between the source and target keypoints following the kinematic tree by aligning the parent bone and then compute the global transformation matrices based on the pre-defined kinematic tree with the forward kinematics. We use the Twist-and-Swing Decomposition \cite{decomp01,hybrik21} to compute each local relative rotation matrix.
We define a bone as the connection between each keypoint and its parent and denote the bone vectors as $\vec s$ for the source and $\vec t$ for the target.
The relative rotation matrix $\mathcal{R}$ between each set of vectors $\vec s$ and $\vec t$ can be formulated as:
\begin{equation}
\label{eq:dec}
\mathcal{R}=\mathcal{R}^{sw} \mathcal{R}^{tw},
\end{equation}
where $\mathcal{R}^{sw}$,$\mathcal{R}^{tw}$represents the swing and twist component of the rotation matrix. The swing rotation has the axis $\vec n$ that is perpendicular to $\vec s$ and $\vec t$ as:
\begin{equation}
\label{eq:perpen}
\vec n=\frac{\vec s \times \vec t}{||\vec s \times \vec t||}.
\end{equation}
$\mathcal{R}^{sw}$ can then be formulated as: 
\begin{equation}
\centering
\begin{aligned}
& \mathcal{R}^{sw} = \textrm{I} +  \sin \alpha[\vec n]_{\times}  + (1-\cos \alpha) [\vec n]_\times^{2}, \\ 
\end{aligned}
\label{eq:sw}
\end{equation}
where $\cos \alpha=\frac{\vec s \cdot \vec t}{||\vec s|| \cdot ||\vec t||}$ with the swing angle $\alpha$. For the twist angle $\phi$, we use a simple network $\mathcal F^{\phi}(\cdot)$ to estimate the $\cos \phi^{s}$ from source and $\cos \phi^{t}$ from target relative to a reference pose:
\begin{equation}
\centering
\begin{aligned}
\cos \phi^{s}=\mathcal F^{\phi}(V^{s}), \quad \cos \phi^{t}=\mathcal F^{\phi}(V^{t}).
\end{aligned}
\label{eq:nettw}
\end{equation}
Subsequently, $\mathcal{R}^{tw}$ can be analytically computed based on the source keypoints skeleton, \ie: 
%
%
\begin{equation}
\centering
\begin{aligned}
\mathcal{R}^{tw} = \textrm{I} + \frac{\sin \phi[\vec s]_{\times}}{||\vec s||^{2}}  +\frac{(1-\cos \phi)}{||\vec s||^{2}} [\vec s]_\times^{2},
\end{aligned}
\label{eq:tw}
\end{equation}
where $\phi=\phi^{t}-\phi^{s}$ represents the relative twist angle. 
$[\vec s]_{\times}$ is the skew-symmetric matrix of $\vec s$. Intuitively, the twist rotation is rotating around $\vec s$ itself, and thus we can determine the twist rotation $\mathcal{R}^{tw}$ according to $\vec s$ and the relative angle $\phi$.
The global transformation matrix for the $k^{th}$ bone $A_{1},\ldots,A_{K}$ can then be analytically computed using forward kinematics with:
\begin{equation}
\centering
\begin{aligned}
{A_{1},\ldots,A_{K}=\mathcal{F}\mathcal{K}(k_{s},\mathcal{R}_{1},\ldots,\mathcal{R}_{K}),}
\end{aligned}
\label{eq:fk}
\end{equation}
where $K$ represents the total number of bones, $\mathcal{F}\mathcal{K}(\cdot)$ represents the whole Forward Kinematics function, $\{\mathcal{R}_1 \ldots, \mathcal{R}_K\}$ represents the relative rotation matrices for all $K$ bones computed from Eqn.~\eqref{eq:dec} and $k_{s}$ represents the source keypoints detected by our keypoints detector from  Eqn.~\eqref{kpnet}. 

The combination of keypoint detection and inverse kinematics (IK) is 
crucial for shape-pose disentanglement. This is because the pose of the target mesh is explicitly extracted as the bone transformations, 
naturally filters out the shape information of the target mesh. Moreover, the transformation matrix, which only depends on the angle between each pair of bone vectors as shown in Eqn.~\eqref{eq:sw}, is invariant to the target scale. 
As a result, we are able to transfer only the pose information from the target to the source while keeping the shape the same. We will show in the experiments that our formulation is better at disentangling the shape and pose information 
in comparison with existing works \cite{disentangle20,limp20,ieugan21} that enforce the disentanglement in an implicit way.


\subsection{Motion Propagation with GMM-based LBS}
With the transformation matrix for all bones, the next step is to propagate the transformations of the sparse bones to all vertices of the source meshes. We use a network $\mathcal F^{S}(\cdot)$ to predict the skinning weights based on the source point cloud and keypoints:
\begin{equation}
\centering
\label{eq:skinpred}
\begin{aligned}
W=\mathcal F^{S}(V_{s},k_{s}),
\end{aligned}
\end{equation}
where $ W \in \mathbb{R}^{N \times K}$. However, given that the ground truth skinning weights are unknown, we design a distance-based method to compute the pseudo skinning weights as supervision. 
Specifically, we model the pseudo skinning weights as a mixture of Gaussians with $K$ bone centers.
The probability of assigning the $i^{th}$ vertex to the $k^{th}$ bone is 
defined as:
\begin{equation}
\centering
\label{eq:gmm}
\begin{aligned}
\bar w_{ik}= \operatorname{Softmax}\biggl( \exp{ \bigl\{-T (v_{i}-C_{k})Q_{k} (v_{i}-C_{k})\bigr\} } \biggr),
\end{aligned}
\end{equation}
where $C_{k} \in \mathbb{R}^3$ is the center of the $k^{th}$ Gaussian, and $Q^b$ is the corresponding precision matrix that determines the orientation and radius of a Gaussian. We only model the radius of the Gaussian, which is predicted by the network with the source as input. $T$ is the hyper-parameter to control the variance of the weights. We use a softmax function to ensure that the probabilities of assigning a vertex to all Gaussian centers sum up to one. Note that for the same identity, we always use the one common pose to compute the pseudo skinning weights based on the observation that the skinning weights for the same identity should not change too much with different poses which could provide a more stable training signal.
We define the skinning weights loss as the $L_{2}$ distance between the network prediction $w_{ik}$ and the pseudo skinning weights $\bar w_{ik}$ as:
%
\begin{equation}
\begin{aligned}
\mathcal{L}_{skin}=\frac{1}{NK}\sum_{i=1}^{N} \sum_{k=1}^{K} (||\bar w_{ik}-w_{ik}||)_{2},
\end{aligned}
\label{eq:skin}
\end{equation}
where $w_{ik}$ is an element of the blend weight matrix $W$, representing how much $k^{th}$ bone transformation affects the vertex $i$. $N$ and $K$ represent the number of vertices and bones. With the skinning weights, the transformation matrix of each vertex can be computed with LBS as:
\begin{equation}
\centering
\label{eq:lbs}
\begin{aligned}
G_{i}&=\sum_{k=1}^{K}w_{ik} A_{k},
\end{aligned}
\end{equation}
where $A_{k}$ is the transformation matrix for the $k^{th}$ bone and $G_{i}$ is the transformation matrix for vertex $i$. Finally, a coarse deformed mesh can be obtained by applying the transformation matrix to each vertex of the source mesh:
\begin{equation}
\begin{aligned}
v^{c}_{i}=G_{i}v^{s}_{i},
\end{aligned}
\label{eq:coarse}
\end{equation}
where $v^{s}_{i}$ and $v^{c}_{i}$ denote the vertices of the source and coarse deformed mesh, respectively.
\subsection{Mesh Refinement}
There are still artifacts in the LBS-based deformed meshes. We further add another refinement network to model the non-linear deformations:
\begin{equation}
\begin{aligned}
\Delta V =\mathcal F^{R}(V_{c},V_{s}), \\
V^{r}=V^{c}+\Delta V.
\end{aligned}
\label{eq:refinement}
\end{equation}
 The input of the refinement network consists of the coarse deformed mesh vertices $V_{c}$ and the source vertices $V_{s}$. $\Delta V$ denotes the predicted deformations and $V_{r}$ the vertices of the refined mesh.
 As shown in Fig.~\ref{fig:framework}, We first extract the point-wise feature of the source shape as the condition and then feed it together with the coarse mesh into the refinement model. More details about the mesh refinement network are provided in the supplementary. Finally, we regard the output mesh of the refinement network as the final deformed mesh where all the loss terms are enforced.
To enforce the shape consistency between the output deformed mesh and the source mesh, we utilize an edge loss for 
shape preservation. Specifically, we enforce the edge length of the deformed mesh to be the same as the input source mesh. This is based on the prior knowledge that all the edges of a mesh should not be stretched or squeezed too much when re-posed. The 
edge loss is defined as the $L_{2}$ distance of the edge length between the source and final deformed meshes:
\begin{equation}
\centering
\begin{aligned}
\mathcal{L}_{edge}=\sum_{i=1}^{N} \sum_{j=1}^{N(i)} (||\bar e_{ij}-e_{ij}||)_{2},
\end{aligned}
\label{eq:edge}
\end{equation}
where 
$\bar e_{ij}$ and $e_{ij}$ represent edges connecting vertex $i$ and $j$ in source and deformed mesh.
\subsection{Cycle and Self Reconstruction}
\label{cylce}
We do not assume the paired ground truth data for training. This means we do not have direct supervision for the output deformed mesh. To further enforce that the pose of the target mesh is transferred to the deformed mesh, we introduce a cycle reconstruction task to reconstruct the input target mesh from the deformed mesh. Specifically, we select three meshes as a triplet: a source mesh ${X_{p1,i1}}$, a target mesh ${X_{p2,i2}}$, and a third mesh with same identity but different pose 
from the target mesh as ${X_{p3,i2}}$. Note that this type of triplet data is easy to obtain since the only requirement is that meshes with the same identity have different poses, which are very common in the existing datasets.
As shown on the right 
of Fig.~\ref{fig:framework}, we first estimate the deformed mesh ${\bar X_{p2,i1}}$ from the source ${X_{p1,i1}}$ and target ${X_{p2,i2}}$ as:
\begin{equation}
\centering
\begin{aligned} 
\bar X_{p2,i1}=F(X_{p1,i1},X_{p2,i2}).
\end{aligned}
\end{equation}
We then use the deformed mesh as the new target mesh and the third mesh ${X_{p3,i2}}$ as the source mesh to reconstruct the original target mesh:
\begin{equation}
\centering
\begin{aligned} 
\bar X_{p2,i2}=F(X_{p3,i2}, \bar X_{p2,i1}).
\end{aligned}
\end{equation}
Intuitively, since the third mesh contains the same identity information as the original target, we can only reconstruct the original target mesh only when the deformed mesh contains the same pose information as the target. Finally, the cycle reconstruction loss is computed as the Point-wise Mesh Euclidean Distance (PMD) between the reconstructed mesh with the target mesh given by:
\begin{equation}
\centering
\begin{aligned} 
\mathcal{L}_{cycle}=\frac{1}{N} \sum_{v=1}^{N}||\bar X_{p2,i2}^{v} -X_{p2,i2}^{v}||_{2}^
{2},
\end{aligned}
\label{eq:cycle}
\end{equation}
where $X^{v}$ represents mesh vertex and $v$ represents index of the vertex. We also adopt the self reconstruction, which transfers pose between the meshes with the same identity, to further enhance the pose transfer. Specifically, we use $X_{p1,i1}$ as the source and $X_{p2,i1}$ as the target to reconstruct $X_{p2,i1}$:
\begin{equation}
\centering
\begin{aligned} 
\bar X_{p2,i1}=F(X_{p1,i1},X_{p2,i1}).
\end{aligned}
\end{equation}
There are two benefits of self reconstruction: 
1) This type of data is easy to obtain, and 2) the supervision can be directly applied to the output deformed meshes.
We minimize the PMD between the reconstructed mesh $\bar X_{p2,i1}$ and $X_{p2,i1}$:
\begin{equation}
\centering
\begin{aligned} 
\mathcal{L}_{self}=\frac{1}{N} \sum_{v=1}^{N}||\bar X_{p2,i1}^{v}-X_{p2,i1}^{v}||_{2}^{2}.
\end{aligned}
\label{eq:self}
\end{equation}
In addition to the point-wise distance for both cycle and self reconstruction, we add an edge length loss between the reconstructed and the original meshes.

\subsection{Total Loss}
The total loss is the weighted summation of all the losses given by:
\begin{equation}
\centering
\begin{aligned} 
\mathcal{L}_{full}=\lambda_{k}\mathcal{L}_{k}+\lambda_{skin}\mathcal{L}_{skin}+\lambda_{cycle}\mathcal{L}_{cycle}\\+\lambda_{self}\mathcal{L}_{self}+\lambda_{edge}\mathcal{L}_{edge},
\end{aligned} 
\label{eq:total}
\end{equation}
where 
$\lambda_{k}$, $\lambda_{skin}$, $\lambda_{cycle}$, $\lambda_{self}$, $\lambda_{edge}$ represent the weights for corresponding loss terms.

\section{Experiments}
\subsection{Datasets}
We conduct our experiments on 4 datasets. NPT~\cite{neuralpt20} is an SMPL-based synthetic dataset that consists of 3D human meshes with different shapes and poses sampled from a specific distribution. We follow the training and testing list used in~\cite{neuralpt20}. SMAL Dataset is a synthetic animal dataset generated with SMAL~\cite{smal17} model. FAUST~\cite{faust14} Dataset consists of real 3D human mesh scans with the same vertices and topology as the SMPL model, which contains 100 meshes including 10 different subjects in 10 poses. Following LIMP, we use the same 80 meshes for training and the remaining 20 meshes for testing.

We also collect a stylized character dataset from Mixamo \cite{mixamo22}, which includes 25 characters and up to 2000 motions for each character. The meshes in this dataset have more complicated shapes, such as humans with clothes and stylized characters. Moreover, the poses of each character are also more diverse, including lying down, squatting, dancing, etc. We use this dataset to validate that our approaches can handle more complicated shapes, poses, and even shapes of different topologies.

\subsection{Implementation Details}
We set the hyper-parameters as $\lambda^{k}=2$, $\lambda^{cycle}=1$, $\lambda^{self}=1$, $\lambda^{edge}=0.0005$, which is the same as NPT for all the datasets. The weights for the skinning weights loss $\lambda^{skin}$ are set differently, namely 0.4 for the NPT and SMAL Dataset and 0.1 for the Mixamo and the FAUST Dataset according to the results of the experiments. We use the multi-stage learning rate decay strategy, where the decay rate of 0.3 is applied for 4 times at the 10,000-th, 20,000-th, 30,000-th, and 40,000-th iterations. Details about our network structure are shown in the supplementary.


\begin{figure}
\centering 
\includegraphics[width=0.48\textwidth]{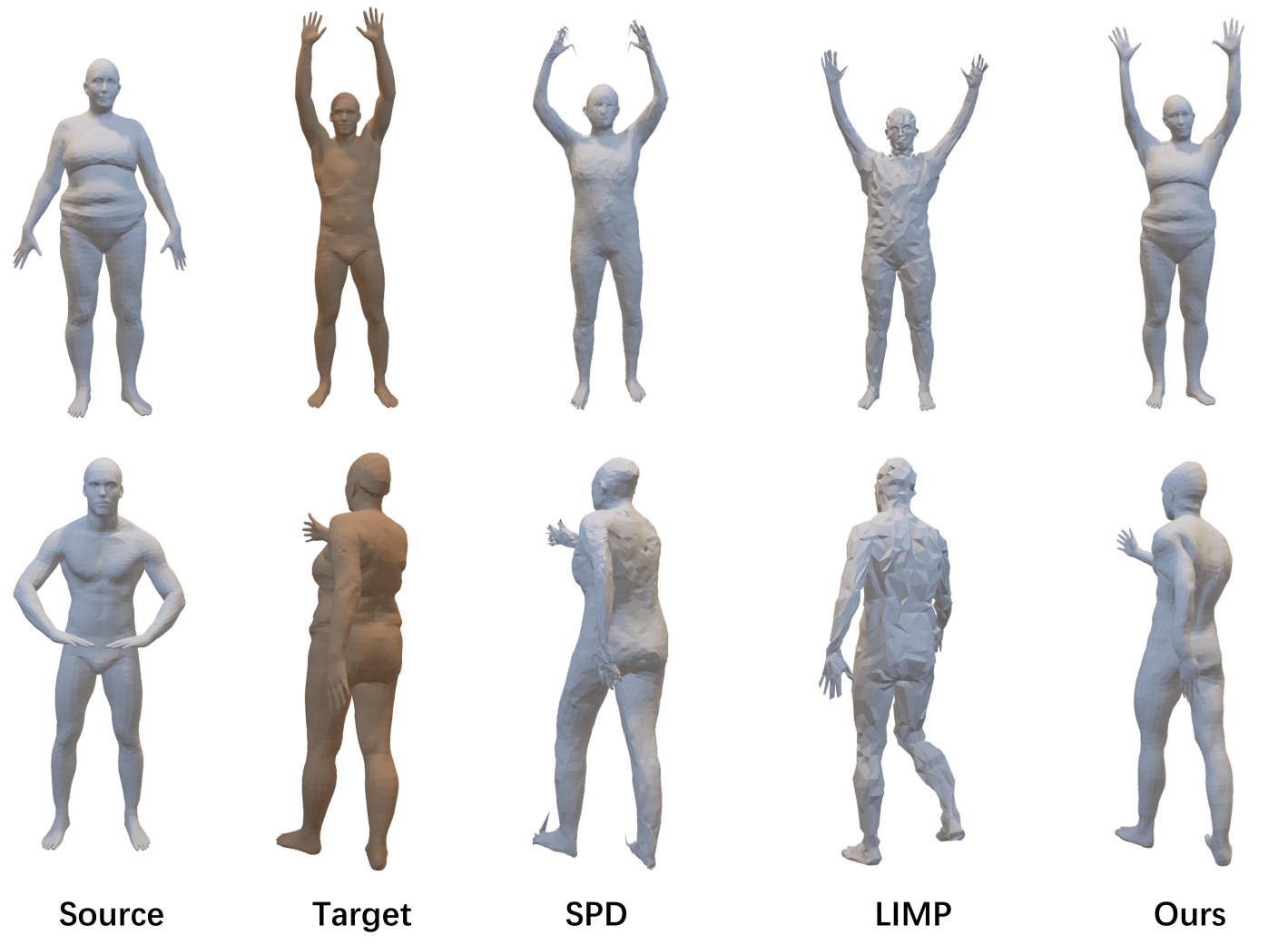} 
\caption{Qualitative comparison on Faust Dataset with LIMP and SPD. 
}
\label{comp1}
\end{figure}

\begin{figure}
  \begin{center}
\includegraphics[width=0.48\textwidth]{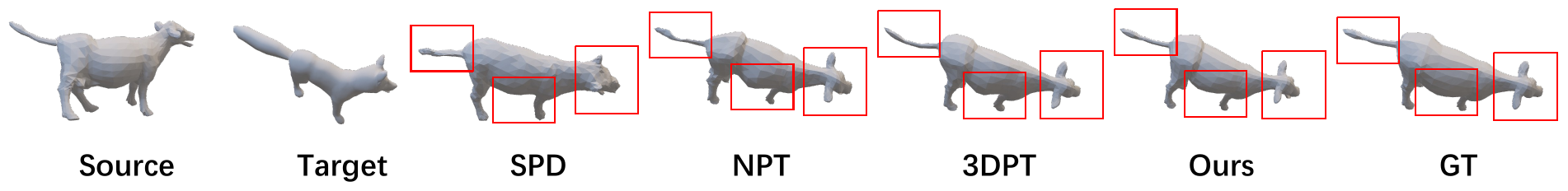}
  \end{center}
  \caption{Pose transfer comparison on SMAL Dataset.}
  \label{fig:smalvis} 
\end{figure}

\begin{figure}
\centering 
\includegraphics[width=0.48\textwidth]{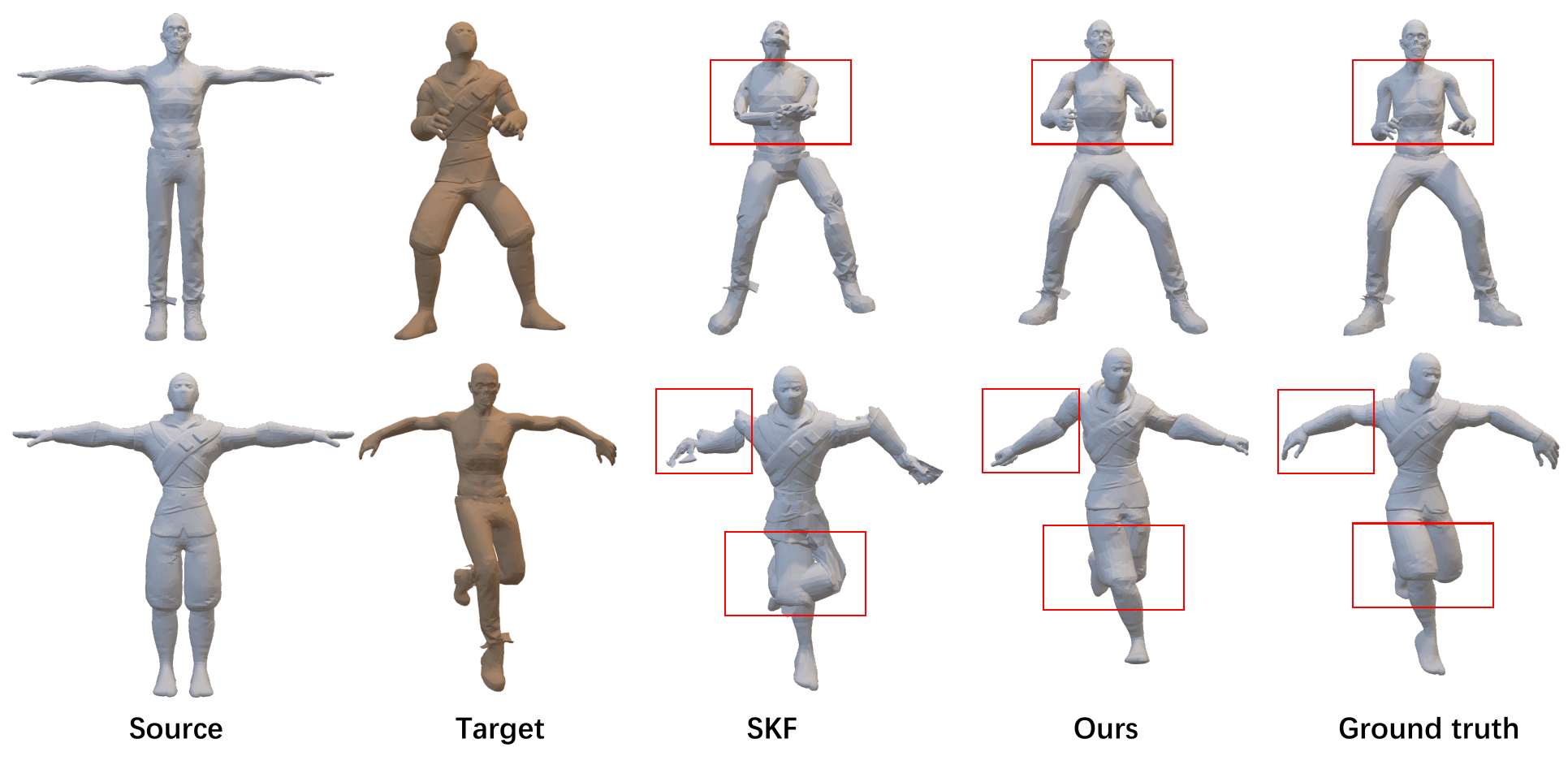} 
\caption{Qualitative comparison on Mixamo Dataset with SKF. 
} \vspace{-3mm}
\label{mixamo}
\end{figure}

\begin{figure}{htbp}
  \begin{center}
\includegraphics[width=0.48\textwidth]{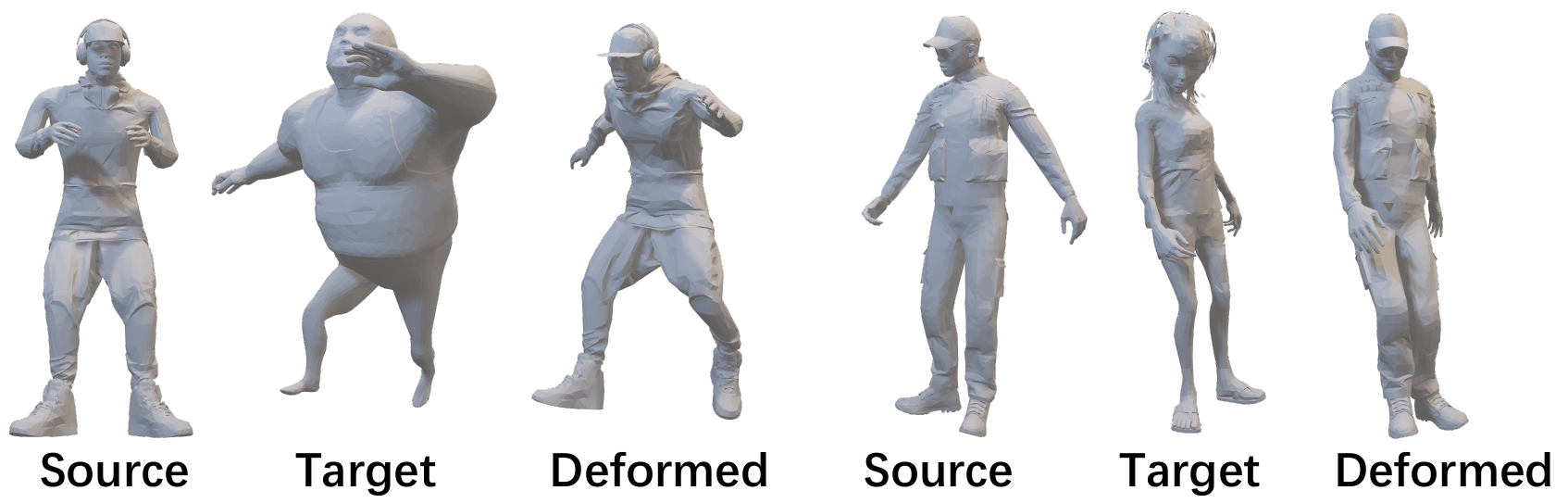}
  \end{center}
  \caption{Cases of larger shape variation.}
  \label{fig:extreme}
\end{figure}

\begin{figure*}[htbp]
\centering 
\includegraphics[width=1\textwidth]{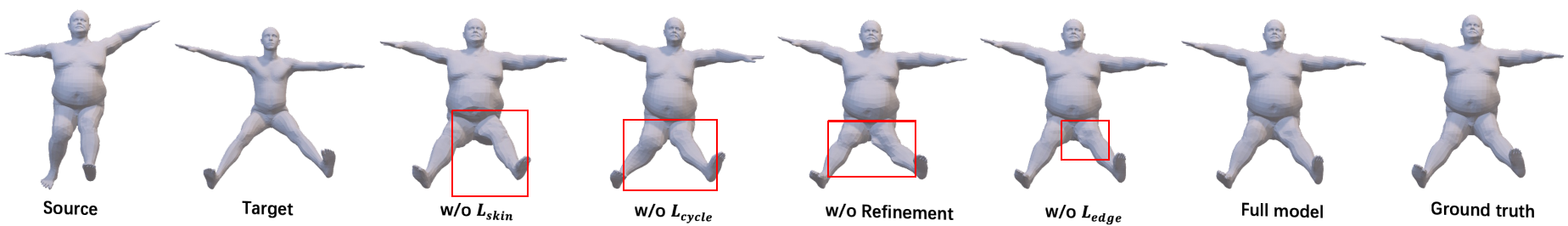} 
\caption{Qualitative comparison of the ablation study on the NPT Dataset. Refer to the text for more details.}
\label{abation}
\end{figure*}  
\subsection{Comparison with Supervised Methods}
Supervised methods can only be trained on synthetic datasets, where ground truth meshes can be synthesized. We first compare our method with the existing supervised approaches on the synthetic dataset using the template (SMPL and SMAL): NPT and SMAL Dataset. The commonly used PMD (1e-4)~\cite{neuralpt20} is used as the evaluation metric. We follow the original train and test split in \cite{neuralpt20} and \cite{3dpt21}. 
The results are shown in Tab.~\ref{npt} and Tab.~\ref{smal}. We can see that our proposed \textit{weakly-supervised} approach achieves comparable or even better performance compared with existing \textit{fully-supervised} approaches on both human and animal datasets. We show qualitative comparison for the animal dataset SMAL in Fig.~\ref{fig:smalvis}. From the figure, we can see our method generate less artifacts compared with the existing methods.
\begin{table}[htbp]
\centering
\setlength{\tabcolsep}{0.15cm}
\begin{tabular}{lccccc}
\hline
 Method& Ours & 3DPT\cite{3dpt21} &GCT\cite{geotrans21} &NPT\cite{neuralpt20}\\
\hline
PMD (1e-4) $\downarrow$ &1.47& \textbf{1.23} &2.3&5.2\\
\hline
\end{tabular} \vspace{2mm}
\caption{Comparison on NPT Dataset with 3DPT, GCT and NPT. Note that all the other methods are \textit{fully-supervised} with ground truth mesh while we are only \textbf{\textit{weakly-supervised}} by keypoints.}
\label{npt}
\end{table}	

\begin{table}[htbp]
\centering
\begin{tabular}{lccccc}
\hline
 Method& Ours & 3DPT\cite{3dpt21} &NPT\cite{neuralpt20}&SPD\cite{disentangle20}\\
\hline
PMD (1e-4) $\downarrow$ &2.98& \textbf{2.26} &6.75 &25.1\\
\hline
\end{tabular} \vspace{2mm}
\caption{Comparison on the SMAL Dataset with 3DPT, NPT and SPD. Note that 3DPT and NPT are \textit{fully-supervised} with ground truth mesh while we are only \textbf{\textit{weakly-supervised}} by keypoints.} \vspace{-2mm}
\label{smal}
\end{table}	
\subsection{Comparison with Unsupervised Methods}
We test our method on the FAUST Dataset and compare it with existing unsupervised approaches. We do not compare with supervised approaches since the paired training data is not available for this dataset. We use the same training and testing dataset with LIMP. The results of LIMP are obtained by directly testing with their pre-trained model. For SPD, we retrain their method on this dataset until convergence since the original model is not trained with the FAUST Dataset. As shown in Tab.~\ref{table:faust}, we can see our method outperforms the unsupervised method SPD, LIMP by a large margin.
\begin{table}[htbp]
\centering
\begin{tabular}{lccccc}
\hline
 Method& Ours & SPD\cite{disentangle20} & LIMP\cite{limp20}\\
\hline
PMD (1e-4) $\downarrow$ & \textbf{11.70}& 22.31&23.51\\
\hline
\end{tabular} \vspace{2mm}
\caption{Comparison on the FAUST Dataset with SPD and LIMP.}
\label{table:faust}
\end{table}
We also show a qualitative comparison on the FAUST Dataset in Fig.~\ref{comp1}. We can see that our results are much better in terms of both shape-preserving and pose transfer. 
The results of LIMP fail to preserve the shape details, shape information of the source mesh is lost in the generated mesh, \eg the belly of the source mesh disappears in their results. Additionally, the pose is not transferred correctly, \eg the pose of the legs shown in the fourth column second row is wrong.
\begin{figure}
  \begin{center}
\includegraphics[width=0.48\textwidth]{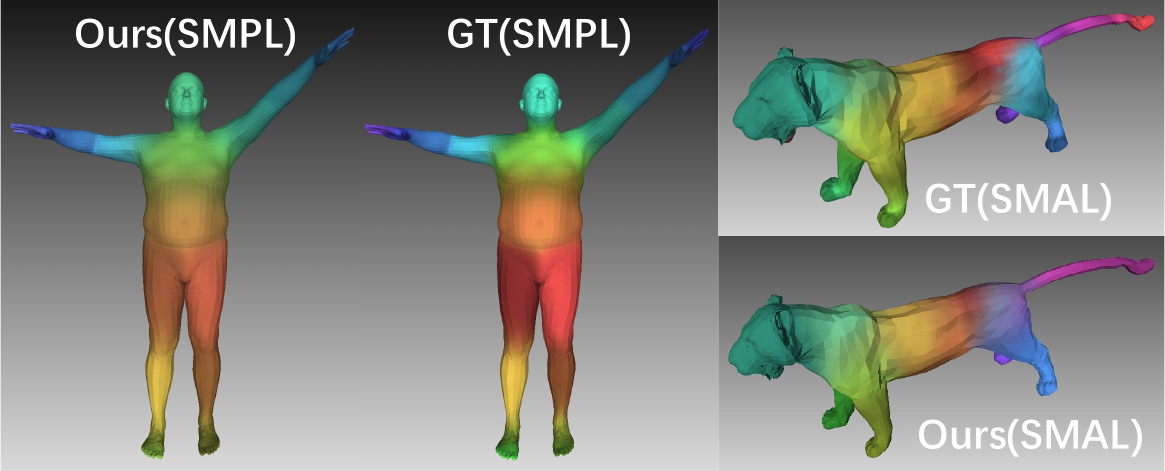}
  \end{center}
  \caption{Skinning weights visualization.}
  \label{fig:skinning}
\end{figure}

For SPD in the third column, both examples show that the generated mesh does not preserve the source shape well. In comparison, our approach can transfer the pose from the target and keep the shape of the source. We credit this good disentanglement to our keypoint-based motion propagation and scalable inverse kinematics which is shape invariant.

\subsection{Cross Topology Evaluation}
\begin{table}[htbp]
\centering
\begin{tabular}{lcccc}
\hline
 Method& Ours&SKF\cite{sk22}\\
\hline
CD (1e-4) $\downarrow$ & \textbf{22.5} & 23.5\\
\hline
\end{tabular} \vspace{2mm}
\caption{Comparison on the Mixamo Dataset with SKF.}
\label{table:com2}
\end{table}	
We further conduct experiments on the Mixamo Dataset to show that our methods can be applied to non-template stylized meshes, \ie meshes with different topologies. We show the results in Tab.~\ref{table:com2}, where we use the Chamfer distance (CD) as the evaluation metric. We compare with SKF \cite{sk22}, which can also handle different topologies. We do not include the supervised approaches 3DPT and NPT since the ground truth is not available for training, we show cross-dataset comparison with them on this dataset in Tab.~\ref{table:cross}.  SPD only works for fixed topology, which is not the case here. We directly take the pre-trained model on the Mixamo Dataset and evaluate it on our test dataset since their test pairs are not available. Note this comparison places our method (Ours) at a disadvantage since SKF is trained with ground truth skinning weights and paired data (when available). Additionally, SKF also requires the T-pose of the driving mesh as input during inference. 
As can be seen from Tab.~\ref{table:com2}, our approach still outperforms SFK although we only use weak keypoint supervision. We also show qualitative comparisons with SKF in Fig.~\ref{mixamo}. As shown in the third column of Fig.~\ref{mixamo}, SKF fails to preserve the geometric details of the source meshes, \eg the unnatural bending at the arms, legs, and hands highlighted in the red box. In comparison, our method successfully transfers the pose from the source to the target and also preserves the shape details of the source mesh. Also, in Fig.~\ref{fig:extreme}, we show our model can transfer pose well in cases with larger shape variation. We show more qualitative results for all the datasets in the supplementary.

\subsection{Cross Dataset Evaluation}
 We show cross-dataset evaluation (PMD $1e^{-3}$) with 3DPT \cite{3dpt21} and NPT\cite{neuralpt20}. 3DPT and NPT are trained on NPT Dataset with ground truth deformed mesh, while ours is weakly-supervised only with keypoints. All methods are tested on the Mixamo Dataset. As shown in Tab~\ref{table:cross}, we can see our method has stronger generalization ability across datasets even with weak supervision.
 \begin{table}[htbp]
\centering
\begin{tabular}{lcccc}
\hline
 Method& Ours&3DPT\cite{3dpt21}&NPT\cite{neuralpt20}\\
\hline
PMD (1e-3) $\downarrow$ & \textbf{15.0} &29.8&23.5\\
\hline
\end{tabular} \vspace{2mm}
\caption{Cross dataset evaluation with 3DPT and NPT.} \vspace{-3mm}
\label{table:cross}
\end{table}	

\subsection{Skinning weights visualization}
We compare our learned skinning weights with ground truth skinning weights on NPT and SMAL datasets. As shown in Fig.~\ref{fig:skinning}, we can see our unsupervised-learned skinning weights are reasonable and similar to ground truth skinning weights.
\subsection{Ablation Study}
We conduct an ablation study for each component of our proposed approach.
As seen from Tab.~\ref{table:ablation}, the error gets larger when each component is removed from the full pipeline, especially for the model without the cycle reconstruction loss or the skinning weights loss. We also show a qualitative comparison in Fig.~\ref{abation} and we highlight the part with obvious artifacts in the red box.
We can see that GMM-based skinning weights loss and cycle reconstruction loss guarantee accurate pose transfer. Without GMM-based skinning weights as supervision, the pose is not transferred well, \eg the left leg is not correct compared with ground truth as shown in the red box. Without cycle reconstruction loss which serves as a pose constraint, the pose of the deformed mesh is also not correct, \eg both legs are in the wrong positions as shown in the red box. 
Refinement network and edge loss help in shape preserving. Without the refinement network, the details on the leg part are not well-preserved. Without the edge loss, the left thigh is stretched too much as shown in the red box. In comparison, both the pose and shape of the deformed mesh from our full model are closer to the ground truth. 
\begin{table}[htbp]
\centering{
\small
\setlength{\tabcolsep}{0.15cm}
\begin{tabular}{ccccc|cc}
\toprule
\multicolumn{5}{c|}{Method} & \multicolumn{2}{c}{PMD (1e-4) $\downarrow$} \\
\cmidrule{1-7}
Refinement& $L_{cycle}$ & $L_{self}$ & $L_{edge}$& $L_{skin}$&  \textit{NPT} & \textit{FAUST}\\
\midrule
\xmark &\cmark & \cmark & \cmark  &\cmark &3.28 &22.3\\
\cmark & \xmark& \cmark & \cmark  &\cmark &6.33 &23.5\\
\cmark & \cmark & \xmark& \cmark  &\cmark &2.44 &18.1\\
\cmark & \cmark & \cmark & \xmark & \cmark &2.40&14.9\\
\cmark & \cmark & \cmark & \cmark & \xmark &6.48&25.6\\
\cmark & \cmark & \cmark & \cmark & \cmark &\textbf{1.47}&\textbf{11.7}\\
\bottomrule
\end{tabular}} \vspace{2mm}
\caption{Ablation studies on the NPT and FAUST Datasets.} \vspace{-2mm}
\label{table:ablation}
\end{table}

\section{Conclusion}
In this paper, we have proposed a novel keypoint-based framework for 3D pose transfer. A cycle reconstruction constraint is designed to enforce self-supervised pose transfer without ground truth. Combining the keypoint-based motion estimation and Scalable IK, our method is able to disentangle shape and pose information better than existing works. In the absence of skinning weights supervision, we design a GMM module to generate pseudo label as guidance. Our approach is topology-agnostic and pose-agnostic, and therefore can be applied to non-template-based 3D meshes with different topologies and source meshes in non-T-pose. Quantitative and qualitative results on several benchmark datasets show the superiority of our proposed approach compared with existing approaches.

\paragraph{Acknowledgement.} This research is supported by the National Research Foundation, Singapore under its AI Singapore Programme (AISG Award No: AISG2-RP-2021-024),
and the Tier 2 grant MOE-T2EP20120-0011 from the Singapore Ministry of Education.

{\small
\bibliographystyle{ieee_fullname}
\bibliography{egbib}

\begin{thebibliography}{10}\itemsep=-1pt

\bibitem{mixamo22}
Adobe.
\newblock Mixamo.
\newblock https://www.mixamo.com, 2022.

\bibitem{decomp01}
Paolo Baerlocher and Ronan Boulic.
\newblock Parametrization and range of motion of the ball-and-socket joint.
\newblock In {\em Deformable Avatars}, 2001.

\bibitem{robo84}
A Balestrino, Giuseppe~De Maria, and L Sciavicco.
\newblock Robust control of robotic manipulators.
\newblock In {\em IFAC Proceedings Volumes}, 1984.

\bibitem{faust14}
Federica Bogo, Javier Romero, Matthew Loper, and Michael~J. Black.
\newblock {FAUST}: Dataset and evaluation for {3D} mesh registration.
\newblock In {\em Computer Vision and Pattern Recognition (CVPR)}, 2014.

\bibitem{leastik05}
Samuel~R. Buss and Jin-Su Kim.
\newblock Selectively damped least squares for inverse kinematics.
\newblock In {\em Journal of Graphics tools}, 2005.

\bibitem{pointnet17}
R.~Qi Charles, Hao Su, Mo Kaichun, and Leonidas~J. Guibas.
\newblock Pointnet: Deep learning on point sets for 3d classification and
  segmentation.
\newblock In {\em Computer Vision and Pattern Recognition (CVPR)}, 2017.

\bibitem{ieugan21}
Haoyu Chen, Hao Tang, Shi Henglin, Wei Peng, Nicu Sebe, and Guoying Zhao.
\newblock Intrinsic-extrinsic preserved gans for unsupervised 3d pose transfer.
\newblock In {\em International Conference on Computer Vision (ICCV)}, 2021.

\bibitem{geotrans21}
Haoyu Chen, Hao Tang, Zitong Yu, Nicu Sebe, and Guoying Zhao.
\newblock Geometry-contrastive transformer for generalized 3d pose transfer.
\newblock In {\em Association for the Advancement of Artificial Intelligence
  (AAAI)}, 2021.

\bibitem{elaimtrans19}
Yugang Chen, Muchun Chen, Chaoyue Song, and Bingbing Ni.
\newblock Cartoonrenderer: An instance-based multi-style cartoon image
  translator.
\newblock In {\em Conference on Multimedia Modeling}, 2019.

\bibitem{limp20}
Luca Cosmo, Antonio Norelli, Oshri Halimi, Ron Kimmel, and Emanuele Rodol{\`a}.
\newblock {LIMP: Learning Latent Shape Representations with Metric Preservation
  Priors}.
\newblock {\em European Conference on Computer Vision (ECCV)}, 2020.

\bibitem{neuralmorph21}
Marvin Eisenberger, David Novotny, Gael Kerchenbaum, Patrick Labatut, Natalia
  Neverova, Daniel Cremers, and Andrea Vedaldi.
\newblock Neuromorph: Unsupervised shape interpolation and correspondence in
  one go.
\newblock {\em Computer Vision and Pattern Recognition (CVPR)}, 2021.

\bibitem{comput85}
Michael Girard and Anthony~A Maciejewski.
\newblock Computational modeling for the computer animation of legged figures.
\newblock In {\em SIGGRAPH}, 1985.

\bibitem{ain17}
Xun Huang and Serge Belongie.
\newblock Arbitrary style transfer in real-time with adaptive instance
  normalization.
\newblock In {\em International Conference on Computer Vision (ICCV)}, 2017.

\bibitem{kd21}
Tomas Jakab, Richard Tucker, Ameesh Makadia, Jiajun Wu, Noah Snavely, and
  Angjoo Kanazawa.
\newblock Keypointdeformer: Unsupervised 3d keypoint discovery for shape
  control.
\newblock In {\em Computer Vision and Pattern Recognition (CVPR)}, 2021.

\bibitem{mvc05}
Tao Ju, Scott Schaefer, and Joe Warren.
\newblock Mean value coordinates for closed triangular meshes.
\newblock In {\em SIGGRAPH)}, 2005.

\bibitem{hybrik21}
Jiefeng Li, Chao Xu, Zhicun Chen, Siyuan Bian, Lixin Yang, and Cewu Lu.
\newblock Hybrik: A hybrid analytical-neural inverse kinematics solution for 3d
  human pose and shape estimation.
\newblock In {\em Computer Vision and Pattern Recognition (CVPR)}, 2021.

\bibitem{sk22}
Zhouyingcheng Liao, Jimei Yang, Jun Saito, Gerard Pons-Moll, and Yang Zhou.
\newblock Skeleton-free pose transfer for stylized 3d characters.
\newblock In {\em European Conference on Computer Vision ({ECCV})}, 2022.

\bibitem{smpl15}
Matthew Loper, Naureen Mahmood, Javier Romero, Gerard Pons-Moll, and Michael~J.
  Black.
\newblock Smpl: a skinned multi-person linear model.
\newblock In {\em International Conference on Computer Graphics and Interactive
  Techniques}, 2015.

\bibitem{handle21}
Liu Minghua, Sung Minhyuk, Mech Radomir, and Su Hao.
\newblock Deepmetahandles: Learning deformation meta-handles of 3d meshes with
  biharmonic coordinates.
\newblock In {\em Computer Vision and Pattern Recognition (CVPR)}, 2021.

\bibitem{3dpt21}
Chaoyue Song, Jiacheng Wei, Ruibo Li, Fayao Liu, and Guosheng Lin.
\newblock 3d pose transfer with correspondence learning and mesh refinement.
\newblock In {\em Neural Information Processing Systems (NeurIPS)}, 2021.

\bibitem{unsupervised22}
Chaoyue Song, Jiacheng Wei, Ruibo Li, Fayao Liu, and Guosheng Lin.
\newblock Unsupervised 3d pose transfer with cross consistency and dual
  reconstruction.
\newblock In {\em IEEE Transactions on Pattern Analysis and Machine
  Intelligence (TPAMI)}, 2023.

\bibitem{meshpro06}
Olga Sorkine.
\newblock Differential representations for mesh processing.
\newblock In {\em Computer Graphics Forum}, 2006.

\bibitem{arap07}
Olga Sorkine and Marc Alexa.
\newblock As-rigid-as-possible surface modeling.
\newblock In {\em Symposium on Geometry processing}, 2007.

\bibitem{dt04}
Robert~W. Sumner and Jovan Popovi{\'c}.
\newblock Deformation transfer for triangle meshes.
\newblock In {\em International Conference on Computer Graphics and Interactive
  Techniques}, 2004.

\bibitem{trans17}
Ashish Vaswani, Noam Shazeer, Niki Parmar, Jakob Uszkoreit, Llion Jones,
  Aidan~N Gomez, \L~ukasz Kaiser, and Illia Polosukhin.
\newblock Attention is all you need.
\newblock In {\em Neural Information Processing Systems (NeurIPS)}, 2017.

\bibitem{neuralpt20}
Jiashun Wang, Chao Wen, Yanwei Fu, Haitao Lin, Tianyun Zou, Xiangyang Xue, and
  Yinda Zhang.
\newblock Neural pose transfer by spatially adaptive instance normalization.
\newblock In {\em Computer Vision and Pattern Recognition (CVPR)}, 2020.

\bibitem{cage20}
Wang Yifan, Noam Aigerman, Vladimir~G. Kim, Siddhartha Chaudhuri, and Olga
  Sorkine-Hornung.
\newblock Neural cages for detail-preserving 3d deformations.
\newblock In {\em Computer Vision and Pattern Recognition (CVPR)}, 2020.

\bibitem{disentangle20}
Keyang Zhou, Bharat~Lal Bhatnagar, and Gerard Pons-Moll.
\newblock Unsupervised shape and pose disentanglement for 3d meshes.
\newblock In {\em European Conference on Computer Vision (ECCV)}, 2020.

\bibitem{smal17}
Silvia Zuffi, Angjoo Kanazawa, David Jacobs, and Michael~J. Black.
\newblock {3D} menagerie: Modeling the {3D} shape and pose of animals.
\newblock In {\em Computer Vision and Pattern Recognition (CVPR)}, 2017.

\end{thebibliography}
}

\end{document}